\documentclass[9pt,twocolumn]{article}

\usepackage[letterpaper,margin=0.45in]{geometry}

\usepackage{amsmath, amssymb, amsfonts}
\usepackage{graphicx}
\usepackage{authblk}
\usepackage{hyperref}
\usepackage{titlesec}
\usepackage{color,soul}
\usepackage{cite}

\titlespacing*{\section}{0pt}{6pt}{3pt}
\titlespacing*{\subsection}{0pt}{4pt}{2pt}

\titleformat{\section}{\bfseries\large}{\thesection}{1em}{}
\titleformat{\subsection}{\bfseries}{\thesubsection}{1em}{}

\begin{document}

\title{\textbf{\large{ YOLOv10 with Kolmogorov-Arnold networks and  vision-language foundation models for interpretable object detection and trustworthy multimodal AI in computer vision perception}}}

\author[1]{Marios Impraimakis\thanks{Corresponding author: Marios Impraimakis (mi595@bath.ac.uk)}}
\author[1]{Daniel Vazquez}
\author[2]{Feiyu Zhou}

\affil[1]{University of Bath, Bath BA2 7AY, UK}

\affil[2]{ Zhejiang University, Hangzhou 310027, China}
\date{}

\maketitle

\begin{abstract}
The interpretable object detection capabilities of a novel Kolmogorov-Arnold network framework are examined here. The approach refers to a key limitation in
computer vision for autonomous vehicles perception, and beyond. These  systems offer limited transparency regarding the reliability of their confidence scores in visually degraded or ambiguous scenes. To address this limitation, a  Kolmogorov-Arnold network is employed as an interpretable post-hoc surrogate to model the trustworthiness of the You Only Look Once (Yolov10) detections using seven geometric and semantic features. The additive spline-based structure of the Kolmogorov-Arnold network enables direct visualisation of each feature’s influence. This produces smooth and transparent functional mappings that reveal when the model’s confidence is well supported and when it is unreliable. Experiments on both Common Objects in Context (COCO), and images from the University of Bath campus demonstrate that the framework accurately identifies low-trust predictions under blur, occlusion, or low texture. This provides actionable insights for filtering, review, or downstream risk mitigation. Furthermore, a bootstrapped language-image (BLIP) foundation model generates descriptive captions of each scene. This tool enables a lightweight multimodal interface without affecting the interpretability layer. The resulting system delivers interpretable object detection with trustworthy confidence estimates. It offers a powerful tool for transparent and practical perception component for autonomous and multimodal artificial intelligence applications.

\end{abstract}

\begin{center}

\end{center}

\section{Introduction}

Object detection systems have attracted interest through the development of the one-stage You Only Look Once models. These detectors achieve real-time performance, making them attractive for deployment in visual perception tasks in autonomous vehicles  \cite{kc2022enhanced, zhong2025yolov8, usama2025vehicle, tang2025using} , traffic \cite{zhang2025psfe}, and pedestrian detection \cite{oussouaddi2025dsr, liu2025vp}.

To this end, Wang et al. \cite{wang2024yolov10} proposed to train the You Only Look Once models without a filtering step for faster and more accurate behavior. Dong et al. \cite{dong2025improved} developed a  method that uses wavelet-based processing, improved feature pooling, and a combined loss design. Luz et al. \cite{da2025smart} showed that combining Internet of Things devices, edge computing, and deep learning enables vehicle detection across hardware with widely varying processing speeds. Wang et al. \cite{wang2025enhanced}  improved the detection of tiny objects by mixing information from different scales and surrounding context \cite{khalili2024sod}. Haoyan et al. \cite{haoyan2025ead} proposed a lightweight model for steel‑surface defect detection by replacing standard layers with an adaptive downsampling structure \cite{akhil2020image, bhatt2021image}. Chhimpa et al. \cite{chhimpa2025transfer}, later, demonstrate that fine-tuning an object detection model can effectively identify brain tumors in medical images. An et al. \cite{an2025research} introduced also a small-target model which improves crack detection using a new head structure, a detail-enhancing pyramid, and a loss function. Zhang et al. \cite{zhang2025vehicle} proposed an improved  network specifically for aerial vehicle detection, Ding et al. \cite{ding2026saa} developed a model that uses attention and multi-scale processing to detect small objects from drone imagery, while Chen et al. \cite{chen2025small} added an edge-focused feature module, improved multi-scale processing, and pruning and distillation steps. Sun et al. \cite{sun2025sod} also combined a transformer-based backbone with an attention-driven feature-fusion method to capture tiny details. Srinivasu et al. \cite{srinivasu2025exploring} studied how adjusting model settings and expanding the training data affects the detection performance. Mahapadi et al. \cite{mahapadi2025real} later optimized two nature-inspired search algorithms. Along these lines, Qin et al. \cite{qin2025acm} proposed a classroom-behavior detection system which improves recognition under crowded and multi-scale conditions. Furthermore, Xie et al. \cite{xie2025yolo} described a model for autonomous-driving detection \cite{wei2025review, li2025investigation, fan2025lud, wei2025scca, li2025road, nie2025investigating, jung2025open} that uses a new gating module, an improved feature pyramid, a redesigned multi-scale network, and a double-distillation training strategy. Liu et al. \cite{liu2025lmg} introduced an enhanced  model designed to locate defects in aircraft-engine components. Finally, Meng et al. \cite{meng2025yolov10} combine ideas from pose-estimation models for agricultural automation.

Toward Kolmogorov-Arnold networks proposed by Liu et al. \cite{liu2024kan}, Shuai et al. \cite{shuai2025physics} introduced later a physics-informed framework that embeds scientific laws into the learning process \cite{adnan2025deep, kiyani2025optimizing,toscano2025kkans,toscano2025aivt}. Wu et al. \cite{wu2025graph} presented Graph Kolmogorov-Arnold networks \cite{li2025kolmogorov}, while Howard et al. \cite{howard2025multifidelity} developed multifidelity Kolmogorov-Arnold networks that use inexpensive low-quality data. Hasan et al. \cite{hasan2025long} combined sequence-learning networks to analyze brain-signal time series, while Baravsin et al. \cite{baravsin2025exploring} conducted a  study across over one hundred time-series datasets for real-world signals. Furthermore, Wang et al. \cite{wang2025intrusion} proposed convolutional Kolmogorov-Arnold networks, which aim to create smaller and more interpretable intrusion-detection models. Saravani et al. \cite{saravani2025predicting} compared Kolmogorov–Arnold networks with  machine-learning approaches, Ansar et al. \cite{ansar2025comparison} modified the loss functions of Kolmogorov-Arnold networks  incorporating a correlation measure to improve performance, while Ta et al. \cite{ta2026fc} introduced fully-composed Kolmogorov-Arnold networks that mix mathematical basis functions  to learn complex relationships from low-dimensional inputs. Finally, Panahi et al. \cite{panahi2025data} proposed a general discovery framework using Kolmogorov-Arnold networks that can uncover governing equations from data.

To the vision-language modeling end, Xue et al. \cite{xue2025blip} introduced a curated dataset, training procedures, and a family of large models that handle both images and text \cite{bowen2026navigating}. Yang et al. \cite{yang2025visionzip} presented VisionZip, a method that keeps only the most useful visual tokens, Xu et al. \cite{xu2025llava} proposed a model with chain-of-thought that performs its own multi-stage reasoning steps, while Vasu et al. \cite{vasu2025fastvlm} introduced a new vision encoder that produces fewer visual tokens to greatly speed up processing. To this end, Deng et al. \cite{deng2025words} studied how vision-language models balance visual and textual information when they are given images and different kinds of text inputs, Zhou et al. \cite{zhou2025vlm4d} introduced the first benchmark fot both space and time in videos, while Shu et al. \cite{shu2025video} proposed video-extra-large which  summarizes the content of each time segment. Deitke et al. \cite{deitke2025molmo} presented some newly created datasets,  Zhan et al. \cite{zhan2025skyeyegpt} also curated a large remote-sensing instruction dataset \cite{li2025lhrs}, and Hu et al. \cite{hu2025rsgpt} created a high-quality remote-sensing image-captioning dataset for aerial images. Furthermore, Wu et al. \cite{wu2025fsvlm} explored  segmenting farmland applications, Dai et al. \cite{dai2025humanvlm} investigated the  interactions between people and their environments, and Xin et al. \cite{xin2025med3dvlm} examined a decomposed three-dimensional encoder, an improved image-text alignment method, and a dual-stream projector for medical images with text descriptions.

Related to interpretable artificial intelligence in engineering \cite{lundberg2020local,joyce2023explainable,jimenez2020drug,novakovsky2023obtaining,lauritsen2020explainable}, Singh et al. \cite{singh2025diaxplain} introduced DiaXplain, a diabetes-diagnosis tool  to deliver understandable results, Roy et al. \cite{roy2025interpretable} developed a  system that emphasizes transparency for clinicians, while Yao et al. \cite{yao2025intelligent} proposed a method that detects interpretable discharge states. Chudasama et al. \cite{chudasama2025toward} introduced TrustKG, a knowledge-graph-based approach that improves the interpretability and reliability of hybrid medical systems. Gliner et al. \cite{gliner2025clinically} demonstrated a tool for electrocardiogram images, Song et al. \cite{song2025interpretable} analyzed climate and lightning data to understand which  conditions increase the likelihood of forest fire-related lightning \cite{ramos2025study}. Rashid et al. \cite{rashid2025ensemble} presented an ensemble-based disease‑detection method, while Salvi et al. \cite{salvi2025explainability} combined uncertainty estimation with interpretability methods. Finally, Agrawal et al. \cite{agrawal2025fostering} explored  explanation techniques to improve the  clarity of model reasoning.

Related to trustworthy artificial intelligence, Chander et al. \cite{chander2025toward} described a range of modern relevant techniques with Wirz et al. \cite{wirz2025re} argued that trust in  outputs depends both on how users perceive the system and on the specific decision situation. Cousineau et al. \cite{cousineau2025trustworthy} considered findings from interviews with developers showing that creating trustworthy artificial-intelligence systems is difficult due to immature standards, inconsistent regulations, unclear definitions, and limited practical tools. Stamboliev et al. \cite{stamboliev2025empty} analyzed European Union policy discourse and concluded that promoting trustworthy artificial intelligence serves as a shared idea that unites diverse political and industry stakeholders. Reinhardt et al. \cite{reinhardt2023trust} examined the philosophical views of trust, Goisauf et al. \cite{goisauf2025trust} provided an interdisciplinary analysis of trust, while Jannel et al. \cite{jannel2026trustability} proposed the concept of trustability as a threshold that distinguishes appropriate trust in these systems. Parra et al. \cite{parra2025lifecycle} introduced the REASON framework, which aims to integrate trustworthiness into the full life-cycle of future communication networks, and Moreno et al. \cite{moreno2025design} proposed a design approach intended to help developers embed trustworthy principles into medical systems.

However, the internal behavior of such models remains difficult to interpret: the learned decision surfaces are highly nonlinear, their dependence on geometric and semantic features is opaque, and their confidence outputs offer limited insight into why detections are accepted, rejected, or assigned particular confidence levels. To this end, a framework that couples You Only Look Once with a Kolmogorov-Arnold networks is examined to model and interpret the internal confidence predictions of the detector. The focus is on the structured seven-dimensional set of interpretable features available at the output of the detection head: normalized bounding-box position, normalized size, predicted confidence, discrete class index, and relative image scale.  Additionally, natural language explanations also offer complementary insight by aligning model outputs with human linguistic reasoning. The combination of these models results in a unified interpretability pipeline: You Only Look Once performs object detection, Kolmogorov-Arnold network provides a transparent surrogate that exposes the structure of the model’s confidence function, and the visual-language model produces natural language descriptions. This multimodal design \cite{song2024multi} brings together visual, numerical, and linguistic interpretability within a single coherent framework.

The remainder of this paper is organized as follows. Section 2 presents the You Only Look Once detector that provides the structured inputs for surrogate modeling. Section 3 introduces the Kolmogorov-Arnold network surrogate, detailing its mathematical foundation, spline-based formulation, and its role in approximating the model's confidence function using interpretable univariate transformations. Section 4 describes the vision-language foundation model component, explaining how natural-language explanations are generated to complement the numerical interpretability offered by the surrogate. Section 5 presents the experimental results, including feature-level analyses, partial-dependence behavior, hidden-unit specialization, edge-importance structure, fidelity evaluation, and qualitative examples demonstrating the interpretability pipeline on the COCO and the University of Bath campus' photos dataset. Section 6 provides a discussion of the interpretability findings, while Section 7 concludes the work.

\section{Modeling confidence by You Only Look Once}

\begin{figure}[hhtbp!]
\centering
\includegraphics[width=8cm]{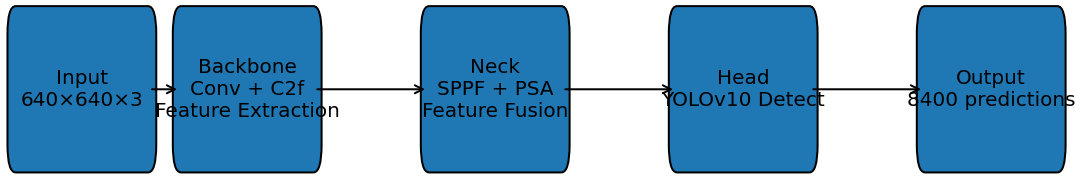}
\caption{Architectural framework of You Only Look Once. C2f stands for coarse-to-fine, SPPF stands for spatial pyramid pooling fast, and PSA stands for  partial self-attention.}
\label{Yolo_arch}
\end{figure}

\begin{figure}[hhtbp!]
\centering
\includegraphics[width=8cm]{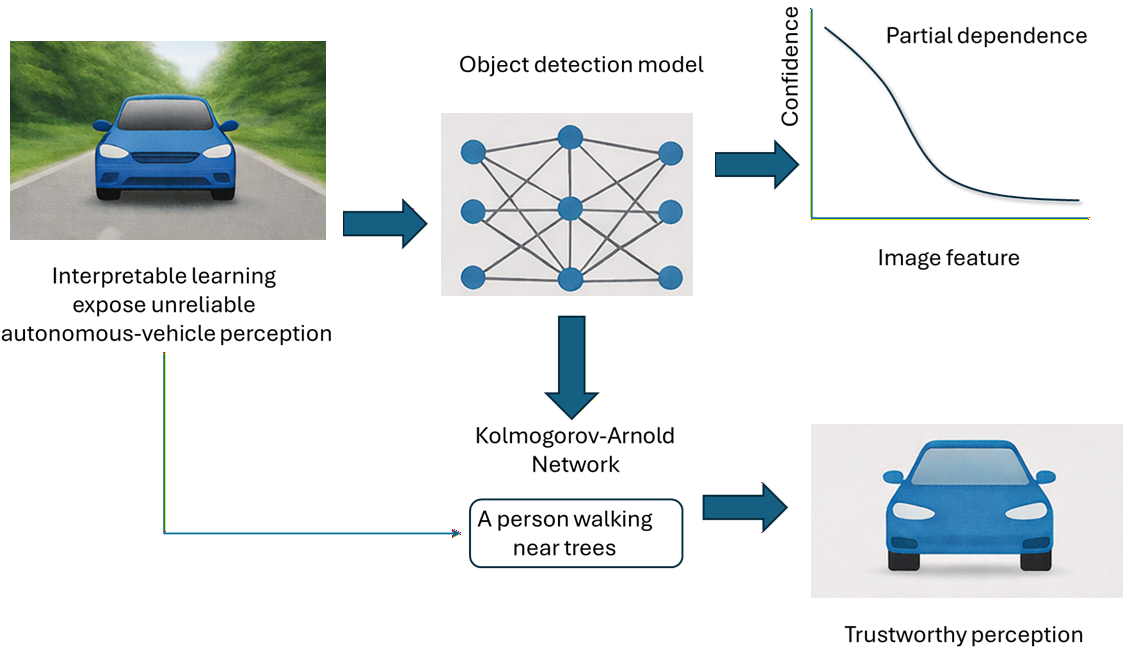}
\caption{Framework for interpretable learning in trustworthy detection perception.}
\label{framework}
\end{figure}

You Only Look Once is a one-stage object detector designed to perform real-time bounding-box prediction that directly regresses object locations and class probabilities seen in Fig. \ref{Yolo_arch}. Given an input image $I \in \mathbb{R}^{H \times W \times 3}$, the model computes a hierarchy of feature maps through a backbone network, producing multiscale feature tensors $[ F_1,F_2,F_3 ]$ that encode semantic and spatial information. At each spatial location on each feature map, the model predicts a bounding box parameterized by its center and dimensions:
\begin{equation}
    b = (x,\, y,\, w,\, h)
\end{equation}
together with a score $o \in [0,1]$ and a distribution over classes $p(c|I)$. The final confidence for a detection is computed as:
\begin{equation}
    \mathrm{conf} = o \cdot p(c|I)
\end{equation}
which serves as the scalar quantity for the Kolmogorov-Arnold Network, Section 3. The network receives an input image of size $640 \times 640 \times 3$, which is processed by the backbone network to extract hierarchical visual features. The backbone primarily consists of convolutional layers and coarse-to-fine modules that progressively learn spatial and semantic representations from the input image. These extracted features are then passed to the neck component to aggregate multi-scale contextual information and enhance feature fusion. The refined feature maps are subsequently forwarded to the detection head, before the model produces a set of dense predictions corresponding to potential object locations and class probabilities across the image, resulting in approximately 8400 candidate detections. 
Let $(\hat{x},\hat{y},\hat{w},\hat{h})$ denote the network’s predicted box and $(x^{*},y^{*},w^{*},h^{*})$ denote the ground truth box assigned to that feature-map location, the model employs a differentiable intersection over union (IoU) metric-based loss function as:
\begin{equation}
    L_{\mathrm{box}}
    = 1 - \mathrm{IoU}\!\left((\hat{x},\hat{y},\hat{w},\hat{h}),\,(x^{*},y^{*},w^{*},h^{*})\right)
\end{equation}
The training objective is expressed as:
\begin{equation}
    L
    = \lambda_{\mathrm{box}} L_{\mathrm{box}}
    + \lambda_{\mathrm{obj}} L_{\mathrm{obj}}
    + \lambda_{\mathrm{cls}} L_{\mathrm{cls}}
\end{equation}
where $\lambda_{\mathrm{box}},\lambda_{\mathrm{obj}},\lambda_{\mathrm{cls}}$ are balancing coefficients. Each detection consists of the tuple $(x,y,w,h,\mathrm{conf},c)$, from which seven numerical features are used by the Kolmogorov-Arnold Network of Section 3, seen as a framework in Fig. \ref{framework}. These are the normalized spatial position $(x,y)$, the normalized size $(w,h)$, the predicted confidence $\mathrm{conf}$, the class index $c$, and the relative image scale $s = \tfrac{wh}{640^{2}}$.

\section{Surrogate Kolmogorov-Arnold network modeling}

\begin{figure}[hhtbp!]
\centering
\includegraphics[width=8cm]{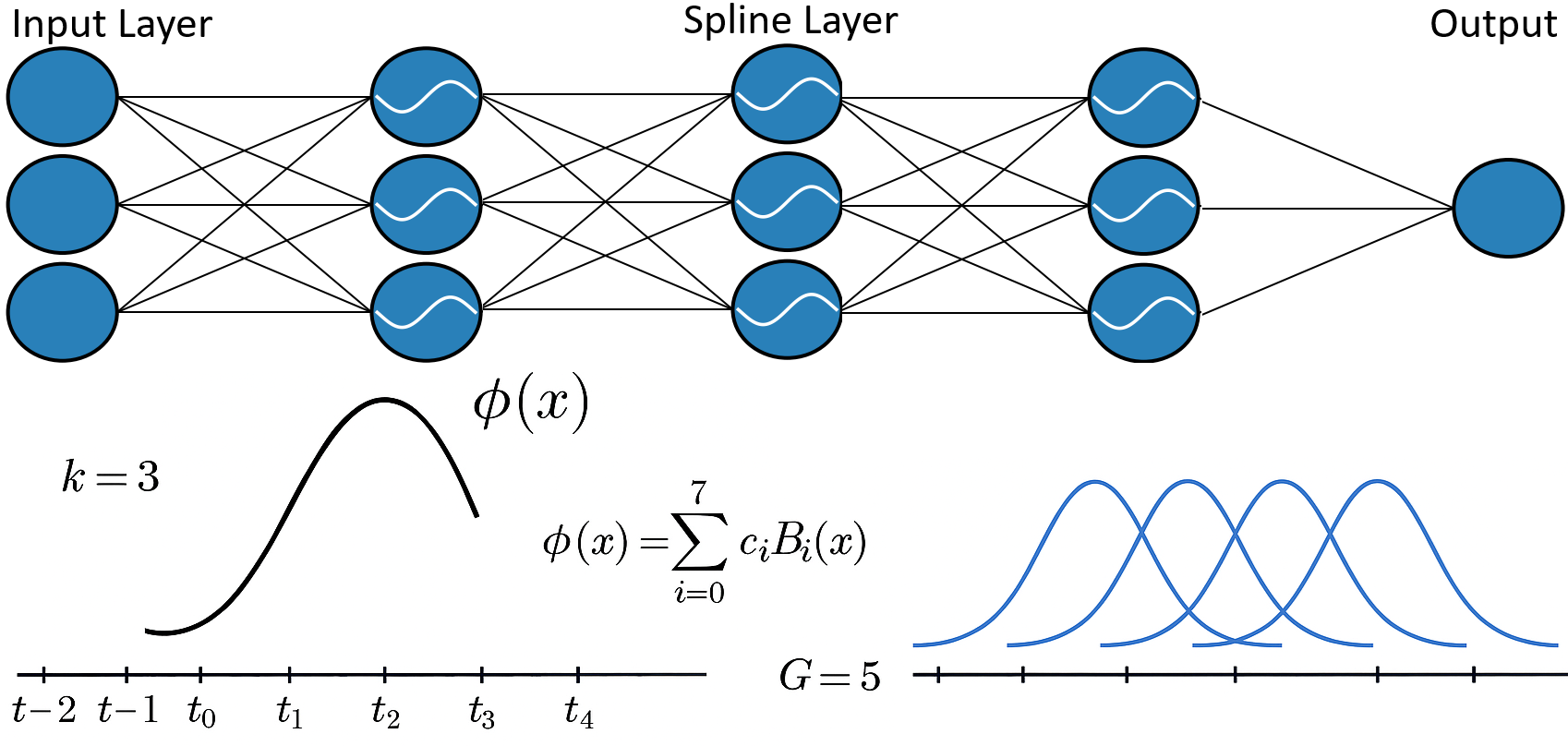}
\caption{Architecture of a Kolmogorov-Arnold network.}
\label{KAN_arch}
\end{figure}

The Kolmogorov-Arnold network  seen in Fig.~\ref{KAN_arch} is inspired by the Kolmogorov-Arnold representation theorem, which states that any multivariate continuous function $f:\mathbb{R}^n \rightarrow \mathbb{R}$ can be expressed as a finite superposition of univariate continuous functions. Kolmogorov proved that there exist continuous functions $\phi_q$ and $\psi_{p,q}$ such that:
\begin{equation}
    f(x_1,\ldots,x_n) = \sum_{q=1}^{2n+1} \phi_q\!\left( \sum_{p=1}^{n} \psi_{p,q}(x_p) \right)
\end{equation}
Kolmogorov–Arnold networks operationalize this idea by replacing the learned $\psi_{p,q}$ functions with trainable spline functions $s_{j,k}(\cdot)$ to every connection between input feature $k$ and hidden unit $j$, and replacing $\phi_q$ with linear combinations of hidden-unit activations as:
\begin{equation}
    h_j(x) \;=\; \sum_{k=1}^{d} s_{j,k}(x_k)
\end{equation}
where $d$ is the input dimensionality. The model output is then obtained as:
\begin{equation}
    y(x) \;=\; \sum_{j=1}^{m} w_j\,h_j(x)
\end{equation}
where $m$ is the number of hidden units and $w_j$ are learned output weights.   The model input signals are 
fed into a hidden layer in which each neuron applies a learnable univariate 
spline function, shown in the diagram as a white spline curve inside each hidden 
unit. The outputs of the hidden spline 
neurons are subsequently combined by a standard output neuron to produce the 
final prediction. 
In this work, the surrogate confidence model is a Kolmogorov-Arnold network with architecture width=[7,16,1], grid=5 and spline order=3 (cubic b-splines), meaning that the network receives seven interpretable numerical features derived from detections. These features consist of normalized bounding–box geometry $(x,y,w,h)$, predicted confidence, discrete class index, and relative image scale $(wh / 640^2)$. Finally,, the model learning form is:
\begin{equation}
    \widehat{c}(x) = \sum_{j=1}^{16} w_j \left( \sum_{k=1}^{7} s_{j,k}(x_k) \right)
\end{equation}
where $x \in \mathbb{R}^7$ is the vector of You Only Look Once-derived features. This architecture directly enables the interpretability results. Partial-dependence plots arise naturally from evaluating a single spline while holding other dimensions fixed. Specifcically, monotonicity emerges from the smoothness properties of the fitted splines, hidden-unit roles can be inferred from correlations between spline outputs and input features, and input-hidden edge importance corresponds to norms of learned spline coefficients. Importanlty, 
each input–hidden connection uses a learnable spline made from B‑spline basis functions with grid = 5 and spline order = 3, each with its own trainable coefficient. During training, these coefficients adjust the curve shape, enabling smooth and interpretable nonlinear mappings while keeping the model simple and stable, seen in Fig.~\ref{KAN_arch}.

\section{Text augmentation by vision-language foundation modeling}

\begin{figure}[hhtbp!]
\centering
\includegraphics[width=8cm]{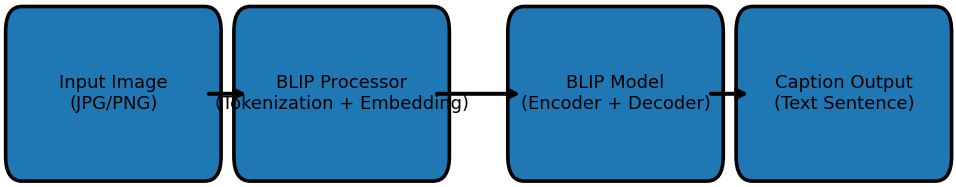}
\caption{Architectural framework of visual-language models. BLIP stands for Bootstrapped Language-Image Pretraining.}
\label{VLM_arch}
\end{figure}

Large vision-language foundation models aim to learn a joint representation between images and natural language, seen in Fig.~\ref{VLM_arch}. An input image $I$ and a textual sequence $T=(t_1,\ldots,t_L)$ are embedded into a unified representation followed by a joint conditioning module:
\begin{equation}
    H = F_{\mathrm{VL}}\bigl(Z_{\mathrm{img}}, Z_{\mathrm{txt}}\bigr)
\end{equation}
where, $F_{\mathrm{VL}}$ consists of stacked transformer blocks, and $Z_{\mathrm{img}}$ and $Z_{\mathrm{txt}}$ are the unified representations of the input. The generative language model then predicts a probability distribution over the vocabulary conditioned jointly on vision and text as:
\begin{equation}
    p(t_{i+1}\,|\,t_1,\ldots,t_i,I) = 
    \mathrm{Softmax}\!\left(
        W\,h_i
    \right)
\end{equation}
where, $h_i$ is the decoder hidden state produced from the fused representation $H$. The input image is first processed by the processor, which performs the required tokenisation, 
normalisation, and embedding transformations. These processed embeddings are then passed to a vision encoder and a text decoder, which jointly generate a natural‑language caption describing 
the visual content of the image. 
Bootstrapped language-image pretraining optimizes a contrastive objective of the form:
\begin{equation}
    L_{\mathrm{ITC}}
    =
    -\log
    \frac{
        \exp\!\left(
            \mathrm{sim}\bigl(Z_{\mathrm{img}},Z_{\mathrm{txt}}\bigr)/\tau
        \right)
    }{
        \sum_{T'} 
        \exp\!\left(
            \mathrm{sim}\bigl(Z_{\mathrm{img}},Z_{\mathrm{txt}}'\bigr)/\tau
        \right)
    }
\end{equation}
where $\mathrm{sim}(\cdot,\cdot)$ is a cosine-similarity operator and $\tau$ is a temperature parameter. The model functions as a language-based interpretability layer that complements the numerical explanations provided by the Kolmogorov-Arnold network of Section 3. For each detection image $I$, BLIP receives the visual input and produces a caption $\hat{T}$ by sampling from the conditional distribution $p(T|I)$ given by the decoder.

\section{Application to vehicle detection tasks}

\begin{figure}[hhtbp!]
\centering
\includegraphics[width=8cm]{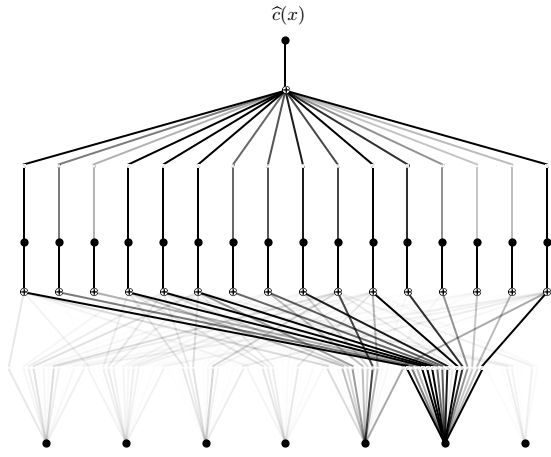}
\caption{Kolmogorov-Arnold network model architecture with seven interpretable inputs and one output. }
\label{fig:kan_arch}
\end{figure}

\begin{figure}[hhtbp!]
\centering
\includegraphics[width=8cm]{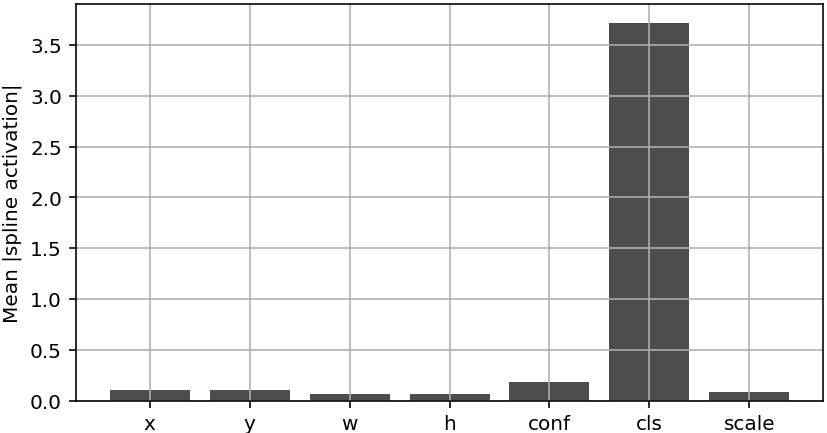}
\caption{Mean absolute spline activation per input feature. }
\label{fig:kan_feature_activation}
\end{figure}

The modeling framework is examined on the Common Objects in COntext dataset focused on  vehicle detection tasks. The goal is to characterize the models internal confidence behavior using the Kolmogorov-Arnold network, and to demonstrate how the spline-based structure provides transparent explanations of the detector’s geometric and semantic sensitivities.  The model, shown in Fig.~\ref{fig:kan_arch}, receives seven input features derived from You Only Look Once detections; bounding-box geometry, predicted confidence, class index, and image scale, and outputs a smooth approximation of the model's confidence. The darker edges in the schematic indicate stronger learned importance. The mean absolute spline activation per input feature is shown in Fig.~\ref{fig:kan_feature_activation}; here, the class index and confidence values dominate the activation magnitudes, while geometric terms $(x,y,w,h)$ exhibit weaker but consistent influence. 

\begin{figure}[hhtbp!]
\centering
\includegraphics[width=8cm]{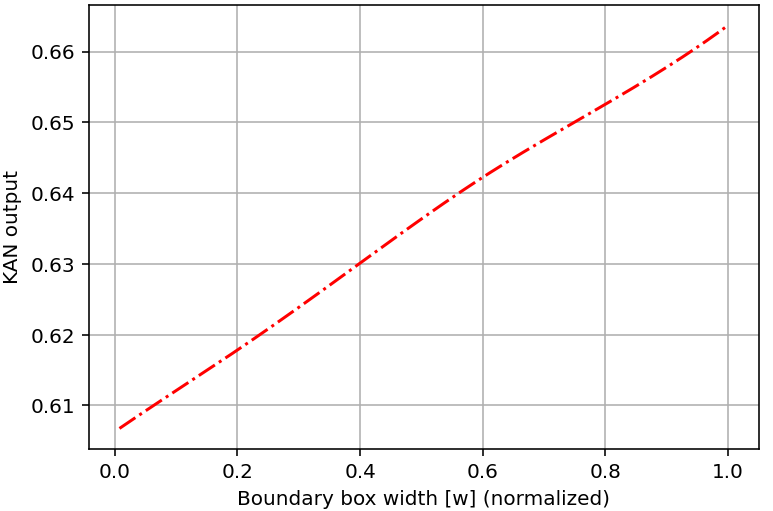}
\caption{Partial dependence of the Kolmogorov-Arnold network on bounding-box width.}
\label{fig:pdp_w}
\end{figure}

\begin{figure}[hhtbp!]
\centering
\includegraphics[width=8cm]{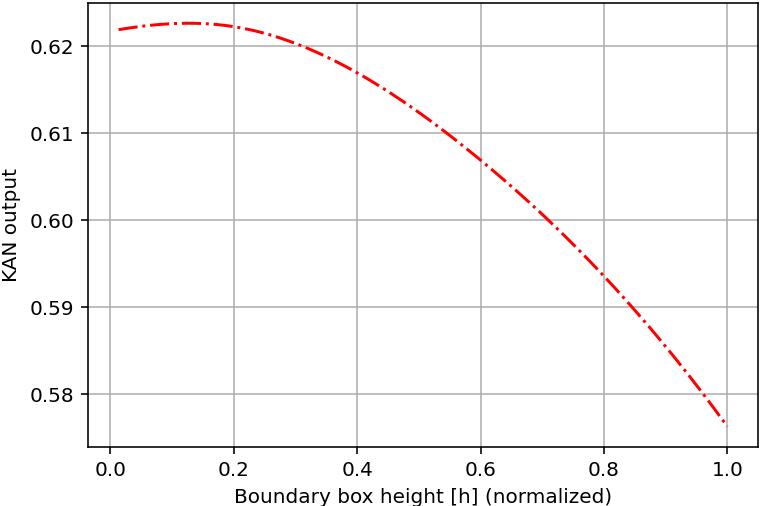}
\caption{Partial dependence on bounding-box height.}
\label{fig:pdp_h}
\end{figure}

\begin{figure}[hhtbp!]
\centering
\includegraphics[width=8cm]{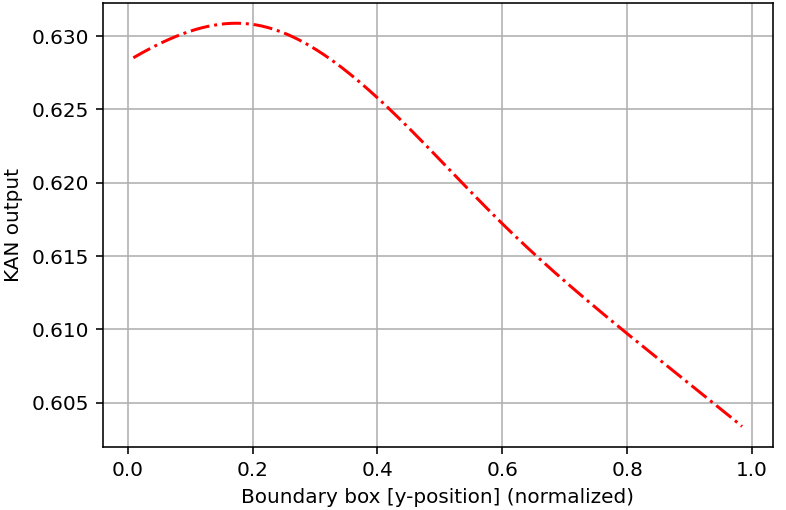}
\caption{Partial dependence on vertical position $y$. }
\label{fig:pdp_y}
\end{figure}

\begin{figure}[hhtbp!]
\centering
\includegraphics[width=8cm]{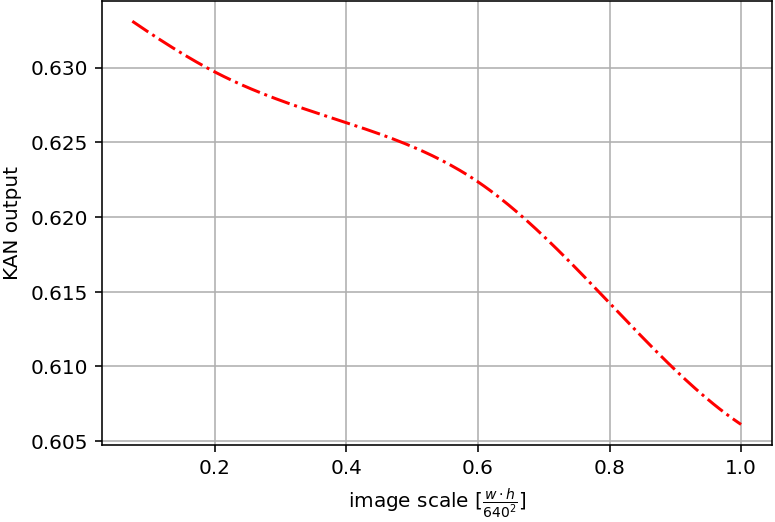}
\caption{Partial dependence on image scale. }
\label{fig:pdp_scale}
\end{figure}

\begin{figure}[hhtbp!]
\centering
\includegraphics[width=8cm]{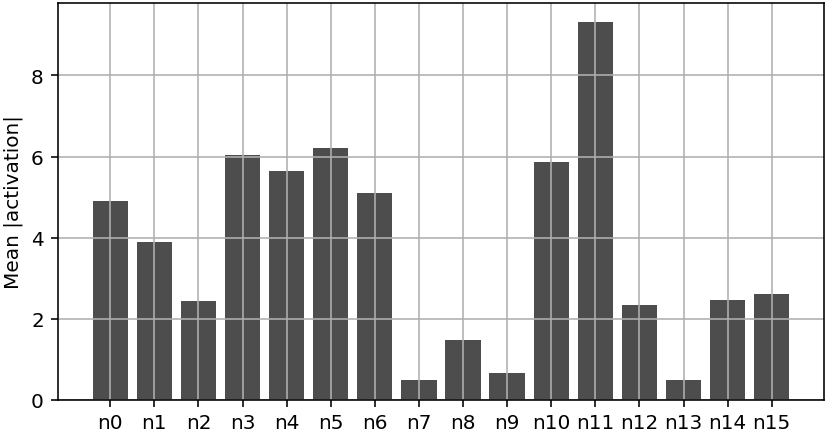}
\caption{Feature influence on a representative hidden unit.}
\label{fig:unit_feature_bars}
\end{figure}

\begin{figure}[hhtbp!]
\centering
\includegraphics[width=8cm]{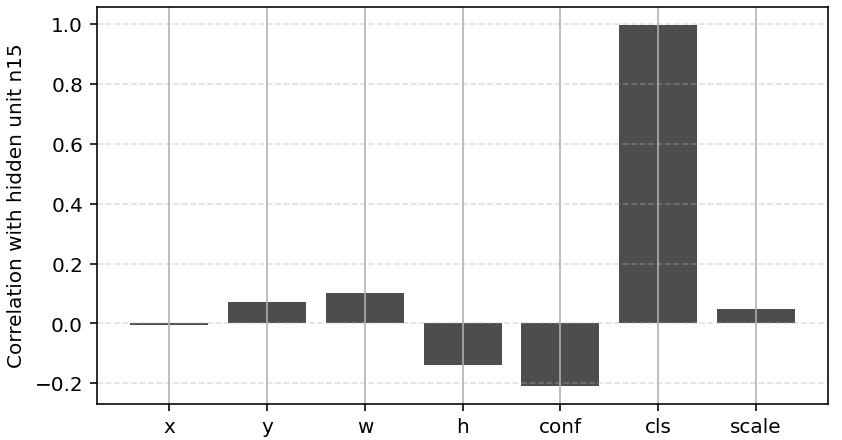}
\caption{Feature influence on a different hidden unit.}
\label{fig:lastunit_feature_bars}
\end{figure}

\begin{figure}[hhtbp!]
\centering
\includegraphics[width=8cm]{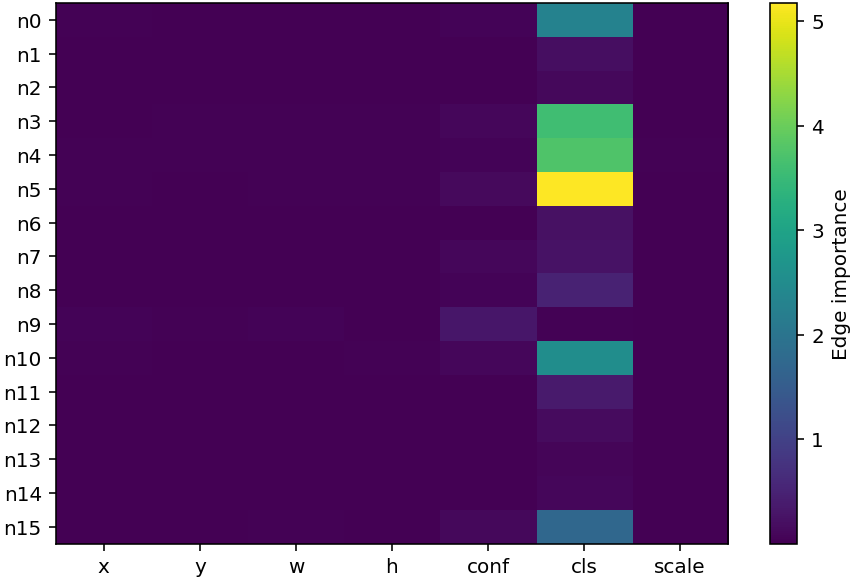}
\caption{Input–hidden edge importance heatmap.}
\label{fig:edge_heatmap}
\end{figure}

\begin{figure}[hhtbp!]
\centering
\includegraphics[width=6cm]{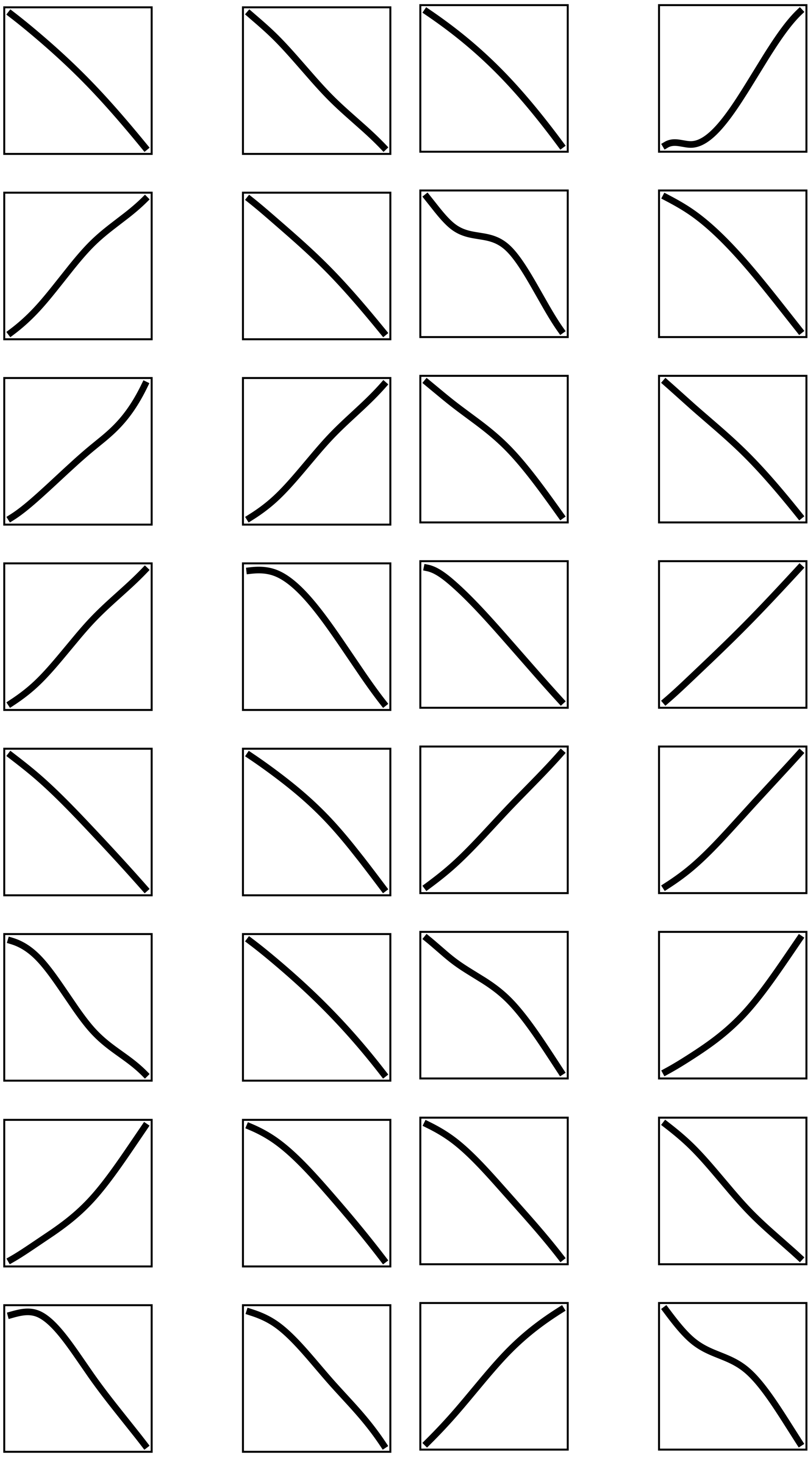}
\caption{Spline functions for all feature - unit pairs (Group 1, horizontal-axis: feature value, vertical‑axis: spline activation).}
\label{fig:all_splines_group1}
\end{figure}

\begin{figure}[hhtbp!]
\centering
\includegraphics[width=6cm]{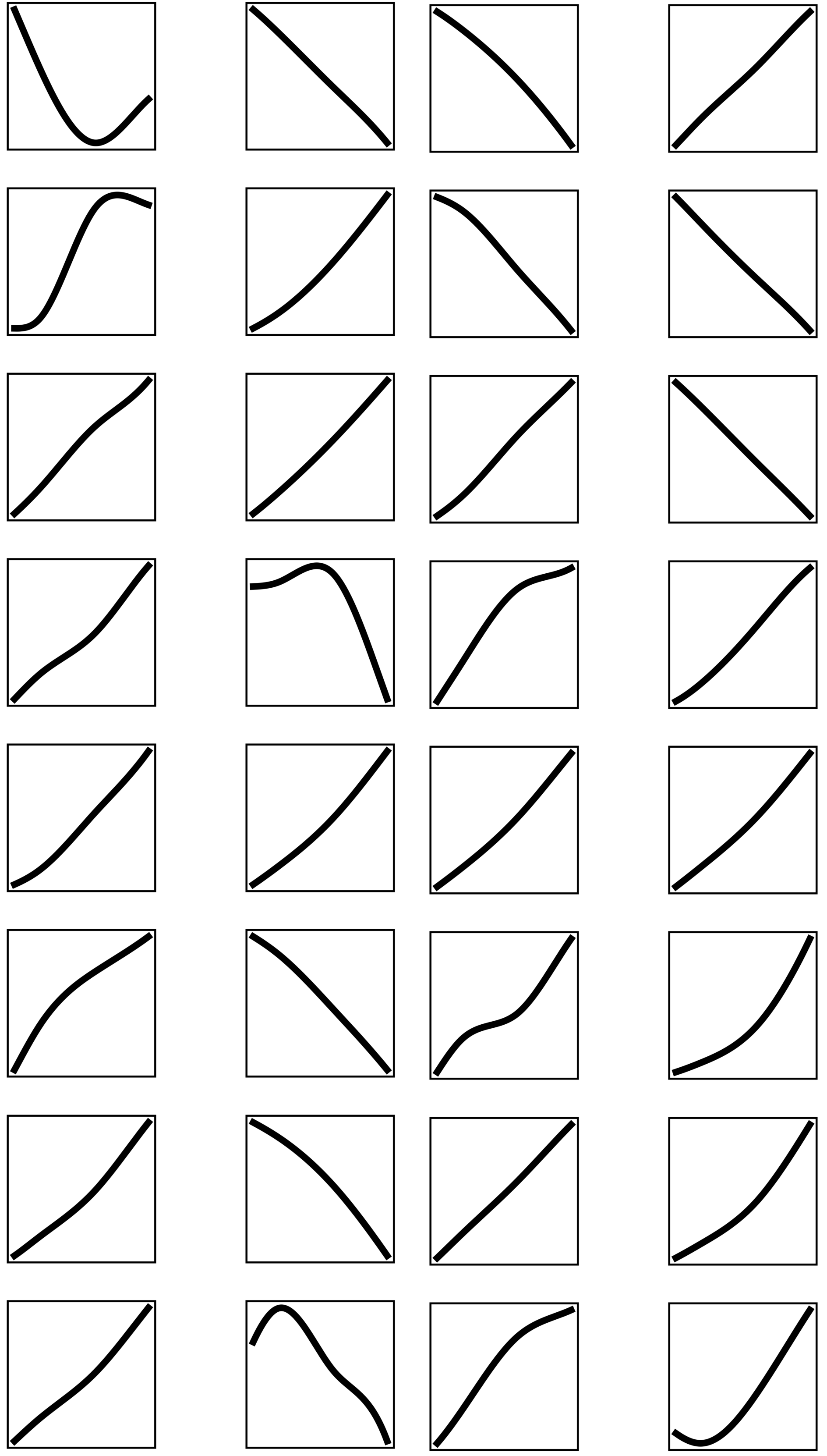}
\caption{Spline functions for all feature - unit pairs (Group 2).}
\label{fig:all_splines_group2}
\end{figure}

\begin{figure}[hhtbp!]
\centering
\includegraphics[width=6cm]{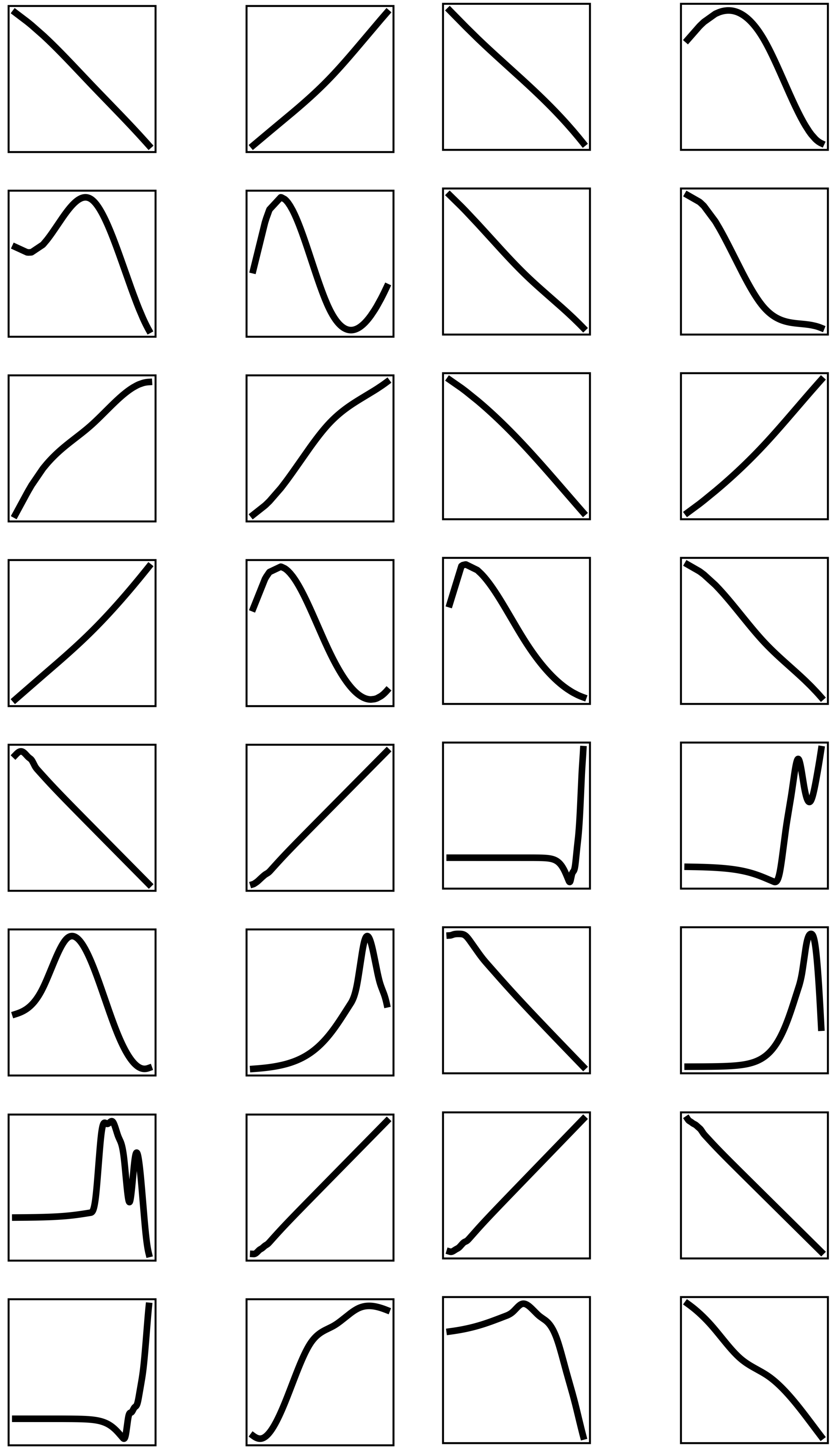}
\caption{Spline functions for all feature - unit pairs (Group 3).}
\label{fig:all_splines_group4}
\end{figure}

\begin{figure}[hhtbp!]
\centering
\includegraphics[width=7cm]{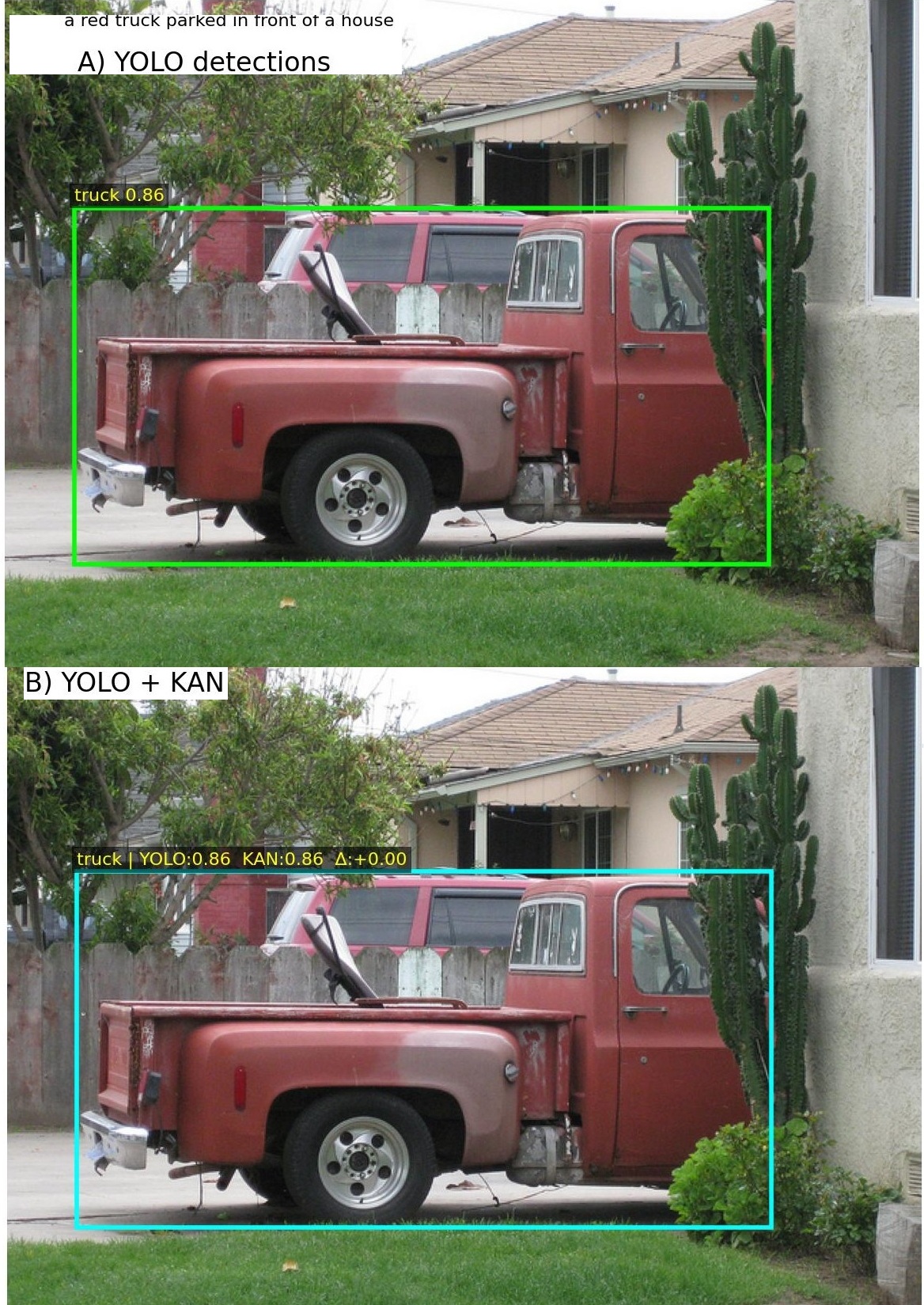}
\caption{Application to a truck scene.}
\label{fig:truck}
\end{figure}

\begin{figure}[hhtbp!]
\centering
\includegraphics[width=7cm]{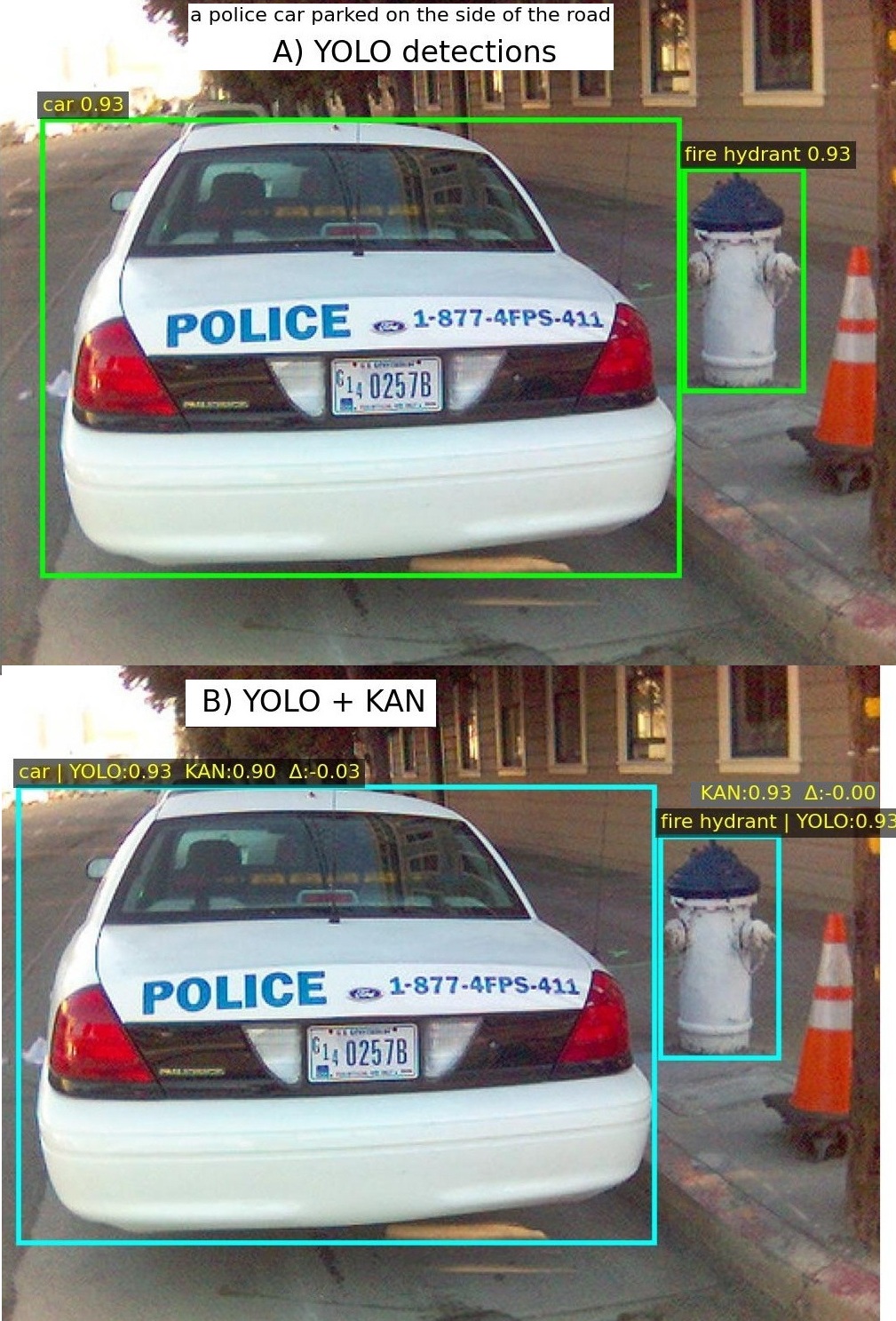}
\caption{Application to a police car.}
\label{fig:police}
\end{figure}

\begin{figure}[hhtbp!]
\centering
\includegraphics[width=7cm]{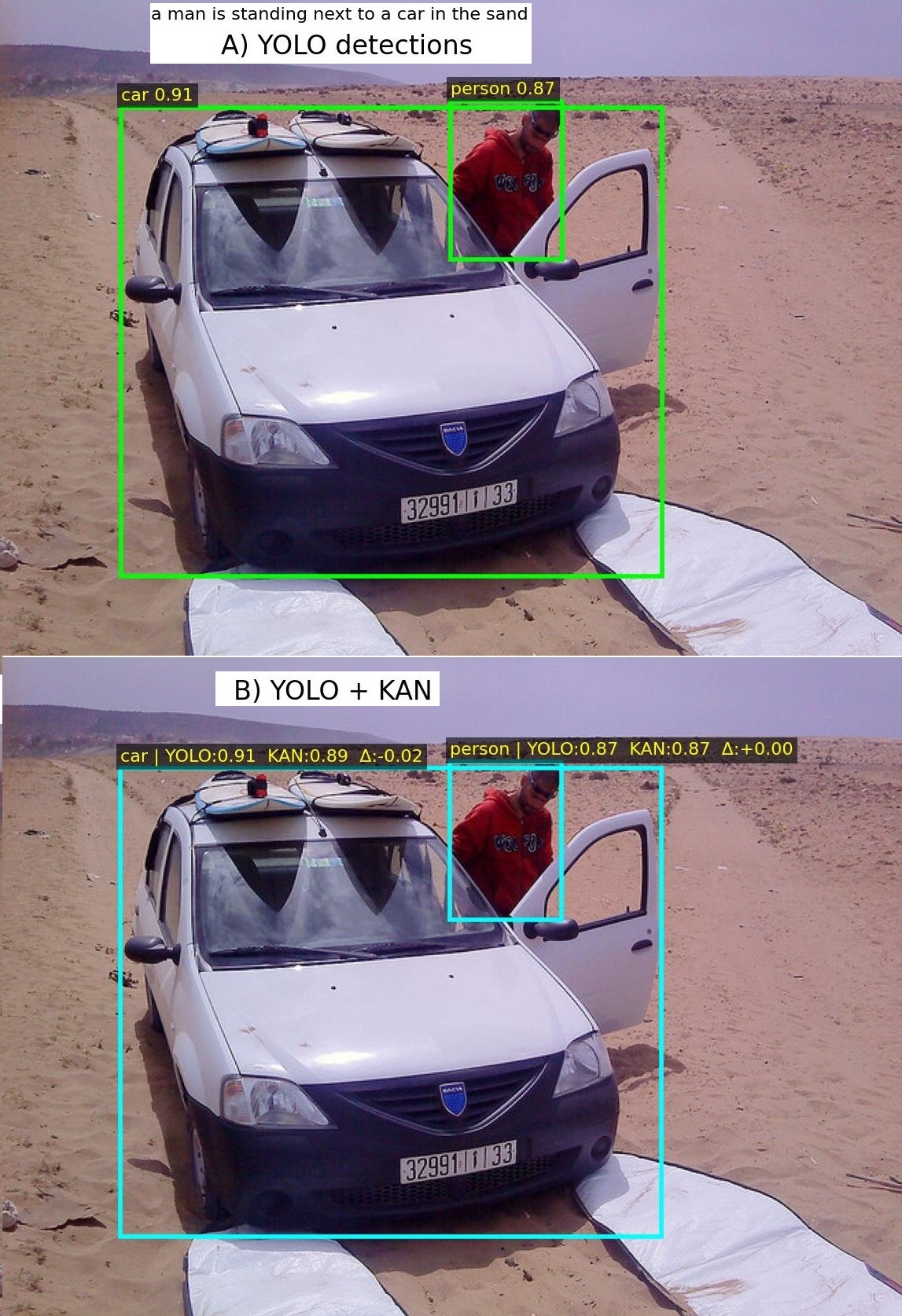}
\caption{Application to a vehicle in sand.}
\label{fig:car_sand}
\end{figure}

\begin{figure}[hhtbp!]
\centering
\includegraphics[width=7cm]{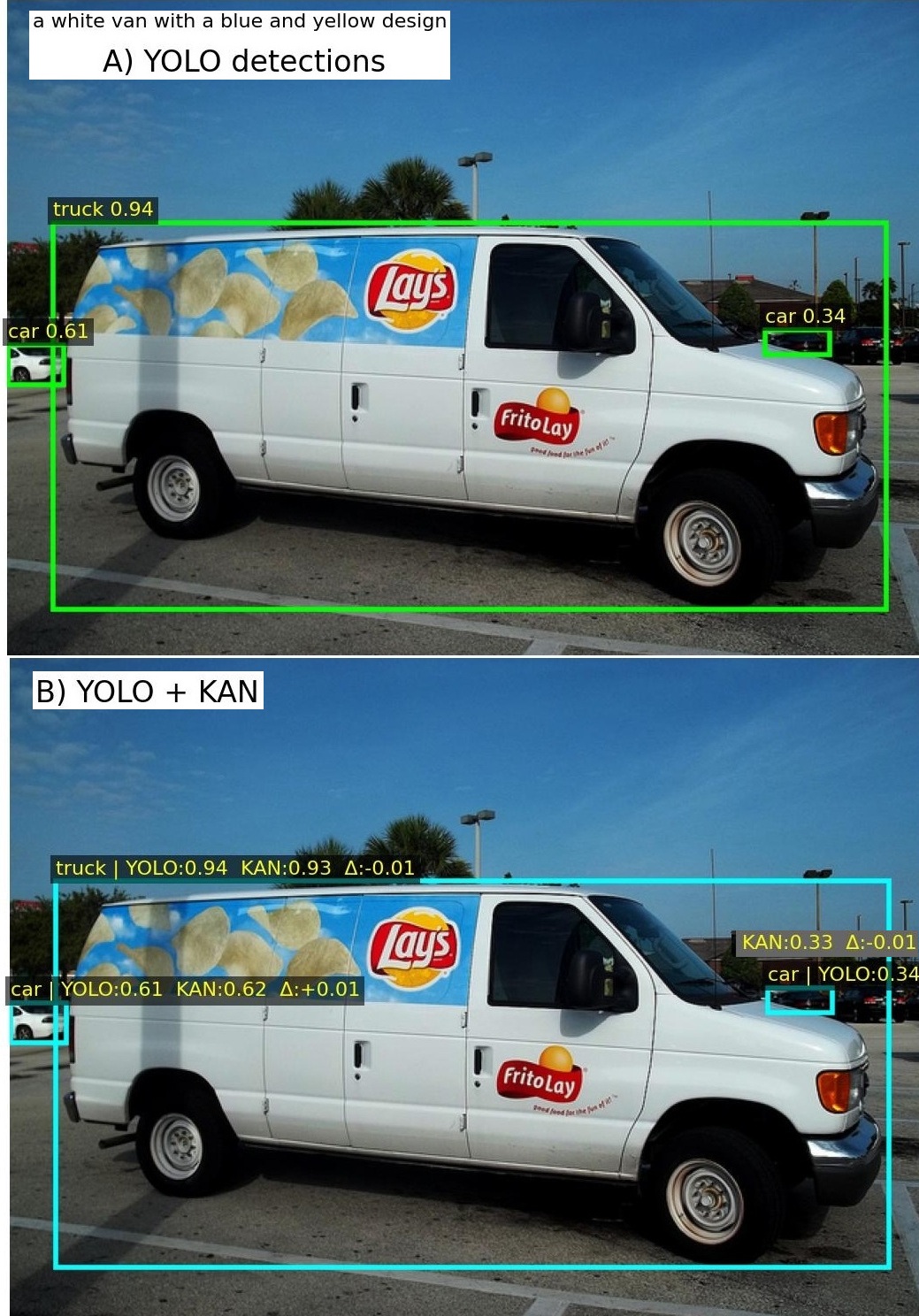}
\caption{Application to a van example.}
\label{fig:truck_lays}
\end{figure}

\begin{figure}[hhtbp!]
\centering
\includegraphics[width=6cm]{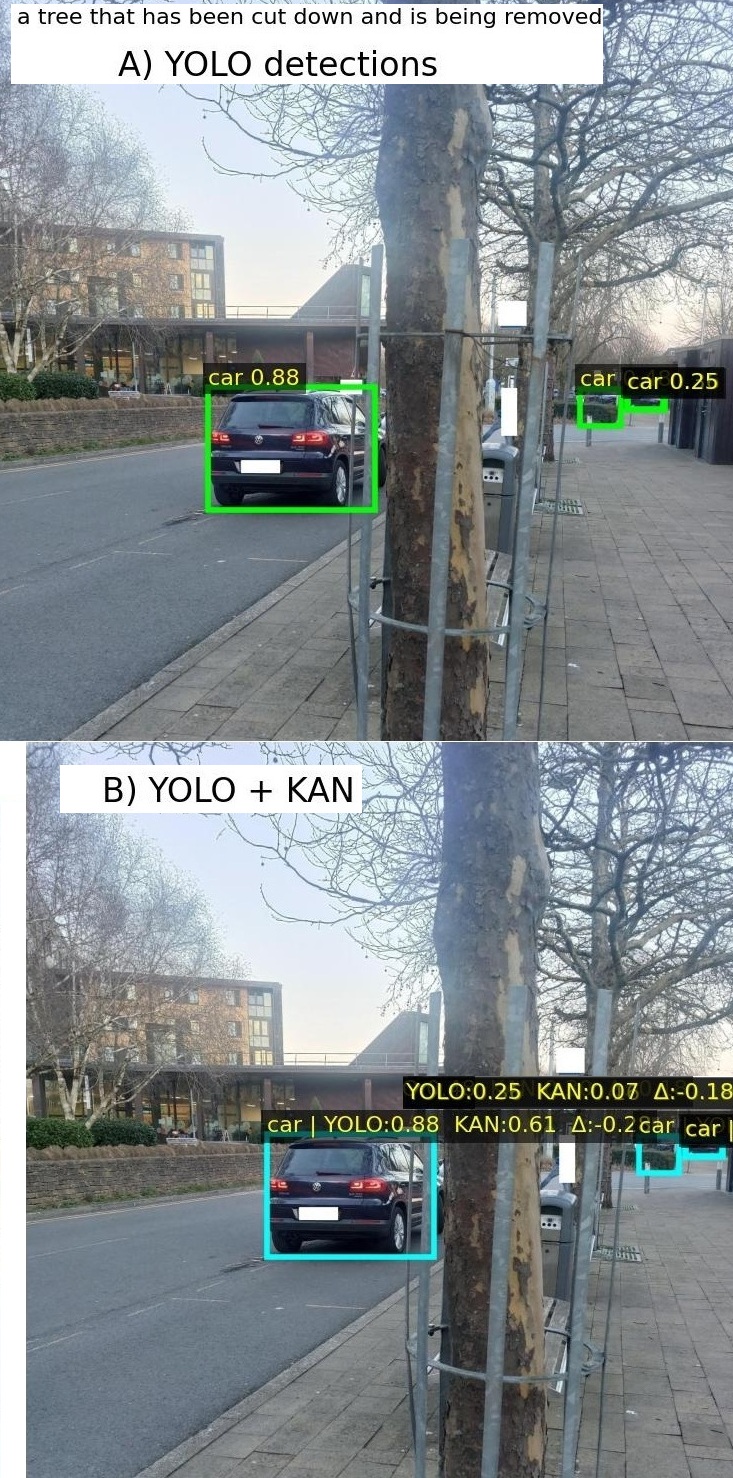}
\caption{Application to a challenging scene with partial occlusion and blur.}
\label{fig:car_tree}
\end{figure}

\begin{figure}[hhtbp!]
\centering
\includegraphics[width=6cm]{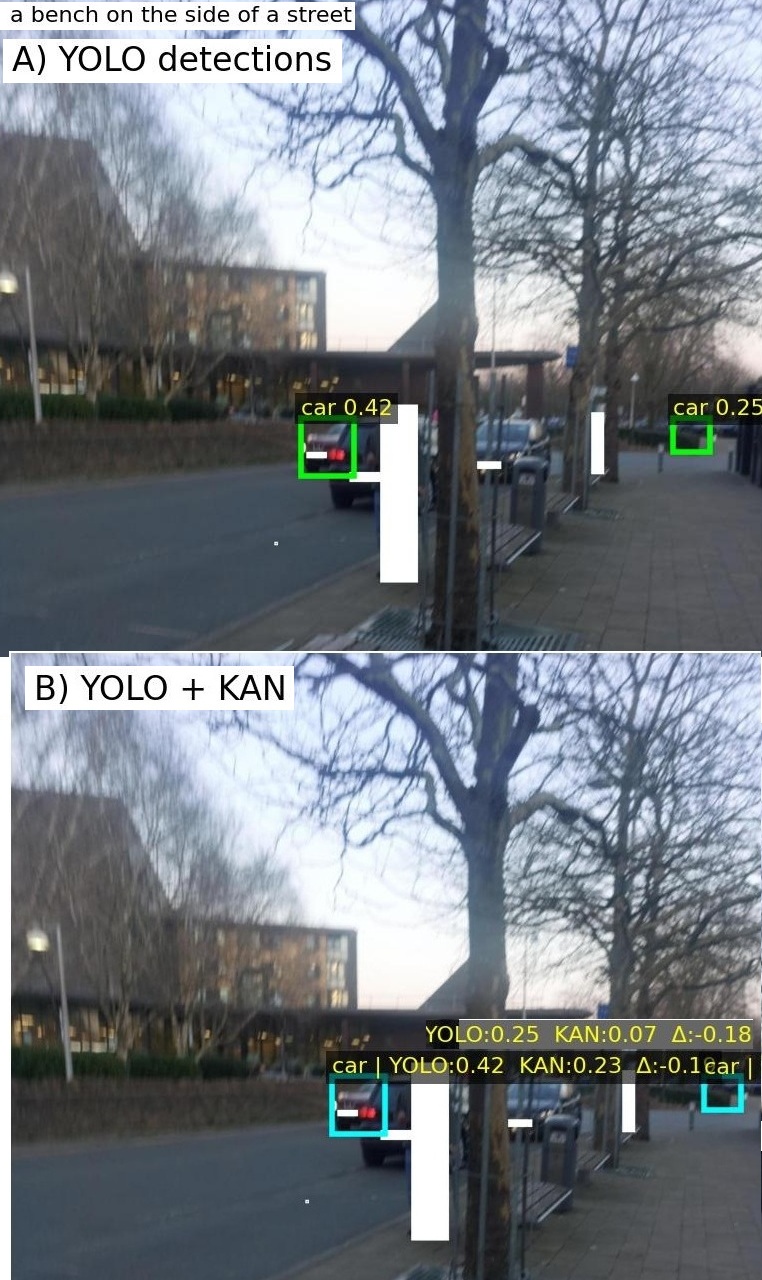}
\caption{Application to a cluttered scene with multiple objects.}
\label{fig:car_bunch}
\end{figure}

\begin{figure}[hhtbp!]
\centering
\includegraphics[width=6cm]{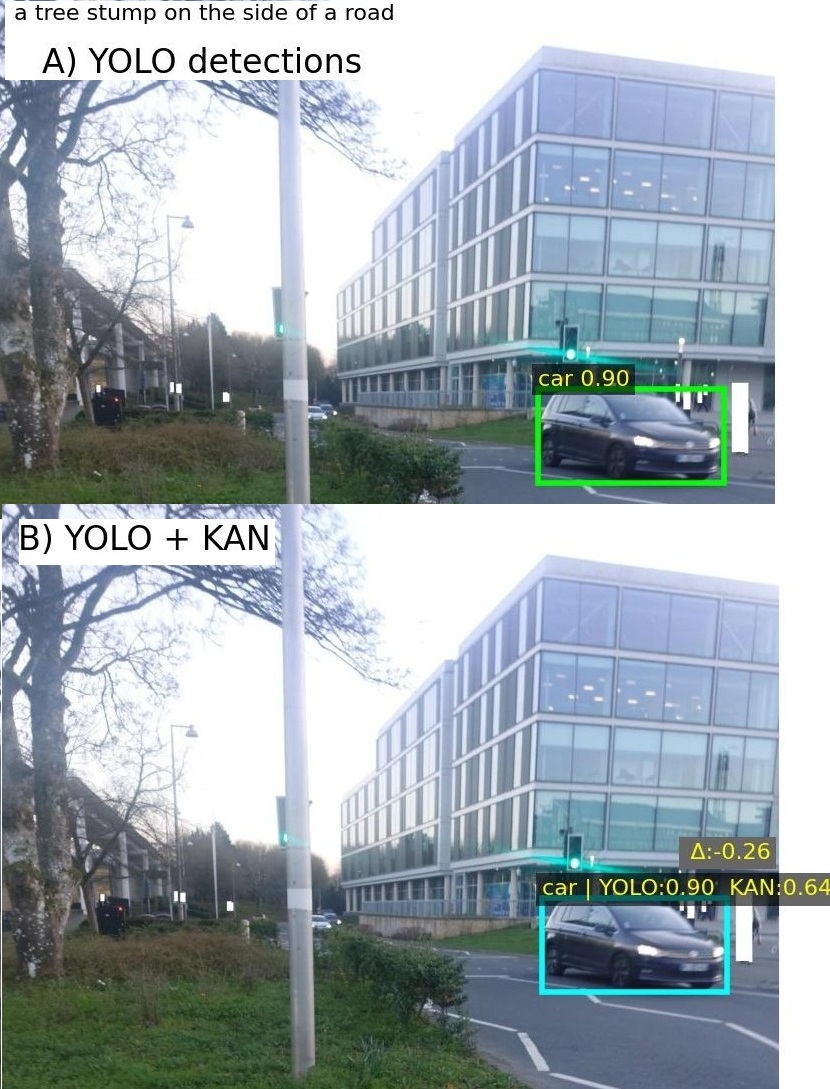}
\caption{Application to a scene with a tree stump and surrounding clutter.}
\label{fig:tree_stump}
\end{figure}

Figures~\ref{fig:pdp_w}-\ref{fig:pdp_scale} present the partial dependence plots for individual features, demonstrating how the Kolmogorov-Arnold network output varies when only one input is changed. The dependence on bounding-box width (Fig.~\ref{fig:pdp_w}) shows a smooth monotonic increase, indicating that the model assigns higher confidence to wider objects. Height (Fig.~\ref{fig:pdp_h}) yields a similarly smooth but slightly weaker trend. The dependence on vertical position $y$ (Fig.~\ref{fig:pdp_y}) is shallow with mild curvature, consistent with weak geometric sensitivity. Image scale (Fig.~\ref{fig:pdp_scale}) exhibits only a small positive effect, confirming that detection confidence is not strongly biased across image resolutions. Across all plots, the curves are smooth and stable, illustrating that the spline components of the  surrogate generalize without overfitting. Figures~\ref{fig:unit_feature_bars} and \ref{fig:lastunit_feature_bars} show the feature-unit correlation bars.  They provide fine-grained interpretability of the internal structure of the surrogate. The full input–hidden edge importance heatmap is also shown in Fig.~\ref{fig:edge_heatmap}, revealing that class and confidence supply strong signals to specific neurons, while geometric inputs connect more weakly.  Figures~\ref{fig:all_splines_group1}-\ref{fig:all_splines_group4} show the collection of spline functions learned by the Kolmogorov-Arnold network for all feature–unit pairs. These visualizations demonstrate the smoothness and diversity of the learned transformations. Each figure corresponds to one group of spline subplots. Finally, Figures~\ref{fig:truck}--\ref{fig:tree_stump} illustrate the application of the full system to real  scenes. You Only Look Once detections are shown alongside Kolmogorov-Arnold network-predicted confidence estimates and vision-language captions. In high-quality images such as the truck (Fig.~\ref{fig:truck}), police car (Fig.~\ref{fig:police}), and sand-car scenes (Fig.~\ref{fig:car_sand}), the surrogate and detector agree closely. For more challenging examples featuring occlusions, blur, or clutter (Figs.~\ref{fig:truck_lays}--\ref{fig:tree_stump}), the Kolmogorov-Arnold network highlights regions where confidence becomes less predictable, demonstrating the surrogate’s ability to expose when detection reliability decreases. The numerical interpretability complements the linguistic explanations generated by the vision-language model.

\begin{table}[h]
\centering
\caption{Feature-level statistics from the network. }
\label{tab:feature_stats}
\begin{tabular}{lccc}
\hline
\textbf{Feature} & \textbf{SplineActivation} & \textbf{Saliency} & \textbf{PD Delta} \\
\hline
x     & 0.107281804 & 1.43$\times10^{-6}$ & $-0.007746$ \\
y     & 0.106276410 & 2.64$\times10^{-6}$ & $-0.025138$ \\
w     & 0.061890736 & 4.17$\times10^{-6}$ & $0.056968$  \\
h     & 0.068972364 & 2.35$\times10^{-6}$ & $-0.045587$ \\
conf  & 0.191350000 & 1.06$\times10^{-4}$ & $0.706939$  \\
cls   & 3.731932000 & 1.67$\times10^{-6}$ & $0.029724$  \\
scale & 0.082707120 & 2.77$\times10^{-6}$ & $-0.0269816$ \\
\hline
\end{tabular}
\end{table}

\begin{table}[h]
\centering

\caption{Hidden-node statistics from the network. }
\label{tab:hidden_nodes}
\begin{tabular}{lcccc}
\hline
\textbf{Node} & \textbf{Activation} & \textbf{Importance} & \textbf{Feature} & \textbf{Correlation} \\
\hline
n0  & 4.908808  & 2.2996383 & cls  & 0.999684334 \\
n1  & 3.8972838 & 0.19116642 & cls & -0.999700487 \\
n2  & 2.4302435 & 0.10419141 & cls & -0.996881783 \\
n3  & 6.031918  & 3.575461   & cls & 0.999689877 \\
n4  & 5.654485  & 3.7371411  & cls & 0.999682188 \\
n5  & 6.2070827 & 5.189023   & cls & 0.999588966 \\
n6  & 5.103179  & 0.22900744 & cls & -0.999275625 \\
n7  & 0.48421517 & 0.27280924 & cls & -0.947130859 \\
n8  & 1.4790272 & 0.48364034 & cls & -0.993801713 \\
n9  & 0.6606282 & 0.2996985  & conf & -0.958717942 \\
n10 & 5.8533688 & 2.5290031  & cls & 0.999058664 \\
n11 & 9.327825  & 0.38113362 & cls & -0.999662876 \\
n12 & 2.3340948 & 0.15058956 & cls & -0.997766674 \\
n13 & 0.4784846 & 0.076826476 & cls & 0.985139787 \\
n14 & 2.45556   & 0.08944808 & cls & -0.999845147 \\
n15 & 2.6013832 & 1.7512177  & cls & 0.997587681 \\
\hline
\end{tabular}
\end{table}

\begin{table*}[h]
\centering
\caption{Input-hidden feature influence values for each Kolmogorov-Arnold network hidden unit. }
\label{tab:edge_importance}
\begin{tabular}{lccccccc}
\hline
\textbf{Node} & \textbf{x} & \textbf{y} & \textbf{w} & \textbf{h} & \textbf{conf} & \textbf{cls} & \textbf{scale} \\
\hline
n0  & 0.029854016 & 0.015423812 & 0.002957478 & 0.014497110 & 0.046715240 & 2.295937500 & 0.005541976 \\
n1  & 0.000579606 & 0.002248535 & 0.001634980 & 0.001924950 & 0.003963163 & 0.191089600 & 0.000456766 \\
n2  & 0.000505880 & 0.002597063 & 0.002021160 & 0.002741036 & 0.005081795 & 0.103393555 & 0.000257480 \\
n3  & 0.011441349 & 0.020771712 & 0.031916957 & 0.032681603 & 0.085905920 & 3.580443100 & 0.004258949 \\
n4  & 0.023302530 & 0.028058290 & 0.026650915 & 0.024421094 & 0.055909153 & 3.746629200 & 0.022205167 \\
n5  & 0.039337154 & 0.018811513 & 0.039905798 & 0.024799882 & 0.136980850 & 5.175523300 & 0.015163474 \\
n6  & 0.003420651 & 0.000805297 & 0.001067619 & 0.001793040 & 0.006243926 & 0.227985070 & 0.002086347 \\
n7  & 0.012705687 & 0.014479717 & 0.004435883 & 0.003144299 & 0.086433806 & 0.245138540 & 0.001393769 \\
n8  & 0.006420781 & 0.012792237 & 0.009923225 & 0.006143517 & 0.056687247 & 0.489789370 & 0.000277586 \\
n9  & 0.045870207 & 0.021616014 & 0.054148242 & 0.003510625 & 0.305599600 & 0.025778690 & 0.019549502 \\
n10 & 0.021460332 & 0.017167496 & 0.017201278 & 0.029365148 & 0.095861495 & 2.544660300 & 0.007906381 \\
n11 & 0.000112305 & 0.002232117 & 0.001292159 & 0.003263123 & 0.007255736 & 0.379950200 & 0.000122972 \\
n12 & 0.002334908 & 0.001076487 & 0.000819806 & 0.001019987 & 0.009017214 & 0.148895730 & 0.000324138 \\
n13 & 0.009948854 & 0.005338817 & 0.003623644 & 0.006526013 & 0.003195604 & 0.074584790 & 0.000295031 \\
n14 & 0.000178230 & 0.001063958 & 0.000743822 & 0.000746206 & 0.000768538 & 0.089053730 & 0.000452965 \\
n15 & 0.019759016 & 0.017285729 & 0.023938041 & 0.010737362 & 0.115140710 & 1.733505800 & 0.001825250 \\
\hline
\end{tabular}
\end{table*}

\begin{table*}[h]
\centering
\caption{Feature influence metrics combining spline activation, saliency, partial dependence delta, and edge importance. }
\label{tab:feature_consistency_reduced}

\begin{tabular}{lccccc}
\hline
\textbf{Feature} &
\textbf{SplineAct} &
\textbf{Saliency} &
\textbf{PDP\_Delta} &
\textbf{EdgeImportance} &
\textbf{Influence} \\
\hline
x     & 0.10728 & 2.86$\times10^{-6}$ & $-0.0077$  & 0.22723  & 0.01739 \\
y     & 0.10628 & 5.29$\times10^{-6}$ & $-0.0251$  & 0.18177  & 0.01392 \\
w     & 0.06189 & 8.34$\times10^{-6}$ & 0.05697    & 0.22228  & 0.04231 \\
h     & 0.06897 & 4.70$\times10^{-6}$ & $-0.0456$  & 0.16731  & 0.00371 \\
conf  & 0.19135 & 2.12$\times10^{-4}$ & 0.70694    & 1.02076  & 0.52001 \\
cls   & 3.73193 & 3.33$\times10^{-6}$ & 0.02972    & 21.05236 & 0.52559 \\
scale & 0.08271 & 5.55$\times10^{-6}$ & $-0.02698$ & 0.08212  & 0.01082 \\
\hline
\end{tabular}
\end{table*}

\begin{table*}[h]
\centering
\caption{Kolmogorov-Arnold network fidelity across overall data and per-feature quantile bins. }
\label{tab:fidelity_bins}
\begin{tabular}{lccccccc}
\hline
\textbf{Scope} & \textbf{Feature} & \textbf{BinIndex} & \textbf{BinRange} & \textbf{N} &
\textbf{R2} & \textbf{MAE} & \textbf{RMSE} \\
\hline
Overall &  &  &  & 9098 & 0.99568467 & 0.010857625 & 0.01510572 \\
Per-Feature & x & 0 & {[}0.0078, 0.2581{]} & 1820 & 0.995543874 & 0.010738888 & 0.014922408 \\
Per-Feature & x & 1 & {[}0.2581, 0.4369{]} & 1819 & 0.995584850 & 0.011135795 & 0.015436348 \\
Per-Feature & x & 2 & {[}0.4369, 0.5569{]} & 1820 & 0.995243065 & 0.011685996 & 0.016162014 \\
Per-Feature & x & 3 & {[}0.5569, 0.7306{]} & 1819 & 0.995711381 & 0.010715223 & 0.015072694 \\
Per-Feature & x & 4 & {[}0.7306, 0.9956{]} & 1820 & 0.996162495 & 0.010012295 & 0.013840625 \\
Per-Feature & y & 0 & {[}0.0090, 0.3730{]} & 1820 & 0.993737923 & 0.012518246 & 0.017162563 \\
Per-Feature & y & 1 & {[}0.3730, 0.4995{]} & 1819 & 0.996216900 & 0.010224011 & 0.014291467 \\
Per-Feature & y & 2 & {[}0.4995, 0.5889{]} & 1820 & 0.996546371 & 0.009600522 & 0.013792516 \\
Per-Feature & y & 3 & {[}0.5889, 0.7164{]} & 1819 & 0.995977398 & 0.010252580 & 0.014582687 \\
Per-Feature & y & 4 & {[}0.7164, 0.9839{]} & 1820 & 0.995056666 & 0.011692083 & 0.015466231 \\
Per-Feature & w & 0 & {[}0.0078, 0.0633{]} & 1820 & 0.994522331 & 0.009916774 & 0.013265161 \\
Per-Feature & w & 1 & {[}0.0633, 0.1175{]} & 1819 & 0.996003601 & 0.009420515 & 0.013088631 \\
Per-Feature & w & 2 & {[}0.1175, 0.1942{]} & 1820 & 0.996316109 & 0.009662537 & 0.013505891 \\
Per-Feature & w & 3 & {[}0.1942, 0.3645{]} & 1819 & 0.995564989 & 0.010814356 & 0.015195336 \\
Per-Feature & w & 4 & {[}0.3645, 1.0000{]} & 1820 & 0.992965135 & 0.014473127 & 0.019501280 \\
Per-Feature & h & 0 & {[}0.0134, 0.0996{]} & 1820 & 0.993620586 & 0.010833444 & 0.014169815 \\
Per-Feature & h & 1 & {[}0.0996, 0.1767{]} & 1819 & 0.994757525 & 0.011014964 & 0.014902389 \\
Per-Feature & h & 2 & {[}0.1767, 0.2982{]} & 1820 & 0.994644570 & 0.011212999 & 0.015743511 \\
Per-Feature & h & 3 & {[}0.2982, 0.5189{]} & 1819 & 0.994923902 & 0.011166411 & 0.015544278 \\
Per-Feature & h & 4 & {[}0.5189, 1.0000{]} & 1820 & 0.995195339 & 0.010060562 & 0.015118543 \\
Per-Feature & conf & 0 & {[}0.2501, 0.3561{]} & 1820 & 0.646103634 & 0.013054217 & 0.018254275 \\
Per-Feature & conf & 1 & {[}0.3561, 0.5248{]} & 1819 & 0.908017482 & 0.010815061 & 0.015045941 \\
Per-Feature & conf & 2 & {[}0.5248, 0.7179{]} & 1820 & 0.953721345 & 0.009243234 & 0.012182652 \\
Per-Feature & conf & 3 & {[}0.7179, 0.8677{]} & 1819 & 0.880090188 & 0.011291197 & 0.014700041 \\
Per-Feature & conf & 4 & {[}0.8677, 0.9854{]} & 1820 & 0.684149733 & 0.009884628 & 0.014724099 \\
Per-Feature & cls & 1 & {[}0.0000, 3.0000{]} & 3558 & 0.999025715 & 0.005320589 & 0.007107449 \\
Per-Feature & cls & 2 & {[}3.0000, 29.0000{]} & 1878 & 0.992935624 & 0.014938466 & 0.019866999 \\
Per-Feature & cls & 3 & {[}29.0000, 52.0000{]} & 1833 & 0.992111145 & 0.014669338 & 0.018721599 \\
Per-Feature & cls & 4 & {[}52.0000, 79.0000{]} & 1829 & 0.994633838 & 0.013618740 & 0.016740484 \\
Per-Feature & scale & 0 & {[}0.0745, 0.6250{]} & 1800 & 0.995370806 & 0.011333028 & 0.015715895 \\
Per-Feature & scale & 1 & {[}0.6250, 0.6672{]} & 1303 & 0.995758532 & 0.010512075 & 0.014713371 \\
Per-Feature & scale & 2 & {[}0.6672, 0.7375{]} & 2343 & 0.995833673 & 0.010658501 & 0.014828051 \\
Per-Feature & scale & 3 & {[}0.7375, 0.7500{]} & 225  & 0.995384340 & 0.011505265 & 0.015818063 \\
Per-Feature & scale & 4 & {[}0.7500, 1.0000{]} & 3427 & 0.995732202 & 0.010832923 & 0.015066167 \\
\hline
\end{tabular}
\end{table*}

\begin{table}[h]
\centering
\caption{Monotonicity analysis of each input feature. }
\label{tab:monotonicity}
\begin{tabular}{lccc}
\hline
\textbf{Feature} & \textbf{MonotonicityScore} & \textbf{Direction} & \textbf{Strength} \\
\hline
x     & 0.021862028  & Flat/Weak  & Weak     \\
y     & 0.034401234  & Flat/Weak  & Weak     \\
w     & 0.376517522  & Positive   & Moderate \\
h     & 0.449024012  & Positive   & Moderate \\
conf  & 0.996658272  & Positive   & Strong   \\
cls   & -0.146067093 & Negative   & Weak     \\
scale & 0.022657916  & Flat/Weak  & Weak     \\
\hline
\end{tabular}
\end{table}

Table~\ref{tab:feature_stats} summarizes the per‑feature statistics extracted from the Kolmogorov-Arnold network. The mean spline activation identifies class and confidence as the dominant drivers of the model’s confidence behavior, while the partial dependence deltas quantify each feature’s global effect across its domain. Geometric inputs play a secondary role, and the uniformly small saliency values demonstrate that the network remains smooth and stable, without sensitivity spikes or overfitting artefacts. Table~\ref{tab:hidden_nodes} provides a detailed description of hidden-unit behavior within the Kolmogorov-Arnold network. Strong correlations between hidden-unit activations and the class feature reveal a consistent specialization pattern, with a few neurons carrying disproportionately large influence over the output. This structure supports a transparent interpretation of the model, where individual units serve identifiable semantic functions rather than mixing signals in an opaque manner. Table~\ref{tab:edge_importance} presents the complete input–hidden edge importance matrix computed from spline coefficients. The results show strong connections from class and, to a lesser extent, confidence, feed into specific hidden units, while geometric features contribute weaker.  Table~\ref{tab:feature_consistency_reduced} consolidates all interpretability metrics into a unified feature-influence score. The strong agreement between normalized statistics confirms that class and confidence dominate the model’s confidence landscape from multiple perspectives; activation strength, global effect, and structural edge contribution. This agreement reinforces the internal coherence of the netowork and provides a robust grounding for the qualitative partial dependence plots analyses. Table~\ref{tab:fidelity_bins} evaluates the Kolmogorov-Arnold network fidelity across quantile bins for each feature. The model maintains  stable fidelity across most of the feature space, demonstrating that the additive spline formulation accurately captures the functional relationship between the model’s confidence and its input parameters. The slight reductions in fidelity for the lowest and highest confidence bins align with the expected difficulty of modeling edge-case detections. Finally, Table~\ref{tab:monotonicity} shows the  monotonicity scores, offering a global characterization of the directionality of the model’s confidence behavior. Confidence emerges as an almost perfectly monotonic driver, providing clear interpretability, while geometric dimensions exhibit moderate monotonic trends. Class shows a weak negative correlation due to index encoding, but combined with the other metrics, its importance remains clearly established.

\section{Discussion}

The interpretability analysis produced by the Kolmogorov-Arnold network reveals several consistent patterns. First, while the full suite of interpretability tables provides detailed quantitative insights, not all tables contribute unique information. The unified feature influence table is the most essential as it captures both the spline-activation behaviour and the structural edge importance values, effectively integrating the core components of two other tables. In contrast, the hidden-unit role analysis is crucial because it exposes how individual units specialize in response to class, confidence, or weaker geometric signals, providing a clear view of how the surrogate decomposes the model's confidence function. The monotonicity scores likewise add interpretive value, showing that the most influential features also tend to exhibit strong or moderate monotonic relationships with the surrogate output. Together, these analyses confirm that the Kolmogorov-Arnold network approximation is not only accurate but also structurally coherent, with meaningful contributions distributed across identifiable units and feature interactions. More specifically, the conceptual diagrams show that the surrogate processes each input feature through dedicated spline transformations, highlighting the interpretability advantage of additive univariate nonlinearities. The feature explanations clarify the semantic meaning of the inputs and confirm that the surrogate assigns the strongest influence to class and confidence, as expected given the nature of the object detection task.

\section{Conclusion}

This work presented a self-awareness framework that enhances the transparency and trustworthiness of You Only Look Once detections through an interpretable Kolmogorov-Arnold network. By modeling confidence as a structured sum of univariate spline functions, the network provides direct insight into how geometric and semantic features shape the reliability of each prediction. The visual-language captioning component assists the perception pipeline with lightweight multimodal descriptions. Overall, the framework provides an interpretable approach with:

\begin{enumerate}
    \item  Numerical insight into whether You Only Look Once's confidence values are trustworthy.
    
    \item Real-time and practical provision of confidence across the full feature domain, while simultaneously identifying unreliable detections in visually ambiguous scenes where the detector may fail.
    
    \item Analysis of hidden units and monotonicity patterns for a coherent internal structure, with specialised neurons that consistently encode class information, confidence trends, and interpretable geometric influences.
    
    \item A multimodal extension using visua-language descriptive scene-level context without altering the underlying modeling.
\end{enumerate}

Importantly, the findings show that the proposed framework enables interpretable  detection with reliable confidence estimation, strengthening the transparency and robustness of modern automated systems.

\section*{Acknowledgements}
The authors gratefully acknowledge the Microsoft Research  (MSR) team for the dataset, while additional data were collected from the University of Bath campus in UK.

\bibliographystyle{unsrt}   

\bibliography{literature}

\end{document}